# An FPGA-based Massively Parallel Neuromorphic Cortex Simulator

Runchun Wang, Chetan Singh Thakur[1], André van Schaik
The MARCS Institute, University of Western Sydney, Sydney, NSW, Australia
mark.wang@westernsydney.edu.au

*Abstract*—This paper presents a massively parallel and scalable neuromorphic cortex simulator designed for simulating large and structurally connected spiking neural networks, such as complex models of various areas of the cortex. The main novelty of this work is the abstraction of a neuromorphic architecture into clusters represented by minicolumns and hypercolumns, analogously to the fundamental structural units observed in neurobiology. Without this approach, simulating large-scale fully connected networks needs prohibitively large memory to store look-up tables for point-to-point connections. Instead, we use a novel architecture, based on the structural connectivity in the neocortex, such that all the required parameters and connections can be stored in on-chip memory. The cortex simulator can be easily reconfigured for simulating different neural networks without any change in hardware structure by programming the memory. A hierarchical communication scheme allows one neuron to have a fan-out of up to 200k neurons. As a proof-of-concept, an implementation on one Altera Stratix V FPGA was able to simulate 20 million to 2.6 billion leaky-integrate-and-fire (LIF) neurons in real time. We verified the system by emulating a simplified auditory cortex (with 100 million neurons). This cortex simulator achieved a low power dissipation of 1.62 µW per neuron. With the advent of commercially available FPGA boards, our system offers an accessible and scalable tool for the design, real-time simulation, and analysis of large-scale spiking neural networks.

## 1. Introduction

Our inability to simulate neural networks in software on a scale comparable to the human brain ($10^{11}$ neurons, $10^{14}$ synapses) is impeding our progress towards understanding the signal processing in large networks in the brain and towards building applications based on that understanding. A small-scale linear approximation of a large spiking neural network will not be capable of providing sufficient information about the global behaviour of such highly nonlinear networks. Hence, in addition to smaller scale systems with detailed software or hardware neural models, it is necessary to develop a hardware architecture that is capable of simulating neural networks comparable to the human brain in terms of scale, with models with an intermediate level of biological detail, that can simulate these networks quickly, preferably in real time to allow interaction between the simulation and the environment. To this end, we are focusing on a hardware friendly architecture for simulating large-scale and structurally connected spiking neural networks using simple leaky integrate-and-fire (LIF) neurons.

Simulating neural networks on computers has been the most popular method for many decades. Software simulators, such as GENESIS (Bower and Beeman, 1998) and NEURON (Hines and Carnevale, 1997), are biologically accurate and model their components with differential equations and sub-millisecond time steps. This approach introduces tremendous computational costs and hence makes it impractical for simulating large-scale neural networks. Other simulators, such as the NeoCortical simulator (NCS; Hoang et al., 2013), Brian (Goodman, 2008) and Neural Simulation Tool (NEST; (Gewaltig and Diesmann, 2007)) are specifically designed for developing large-scale spiking neural networks. Because these tools perform numerical simulations, they do not scale very well, slowing down considerably for large networks with large numbers of variables. For instance, the Blue Gene rack, a two-million-dollar, 2048-processor supercomputer, takes one hour and twenty minutes to simulate one second of neural activity in 8 million integrate-and-fire neurons connected by 4 billion static synapses (L. Wittie, H. Memelli, 2010). Graphic Processing Units (GPUs) can perform certain types of simulations tens of times faster than a PC (Shi et al., 2015). However, as GPUs are still performing numeric simulations, it can take hours to simulate one second of activity in a tiny piece of cortex (Izhikevich and Edelman, 2008).

Along with general hardware solutions, there have been a number of large-scale neuromorphic platforms such as Neurogrid (Benjamin et al., 2014), BrainScaleS (Pfeil et al., 2013), TrueNorth (Merolla et al., 2014), SpiNNaker (Furber et al., 2014) and HiAER-IFAT (Park et al., 2016). In Neurogrid, sub-threshold analogue circuits are used to model neuron and synapse dynamics in biological real time, with digital spike communication. A 16-chip board is capable of simulating one million neurons, using a two-compartment model with ion-channels, in real time. In BrainScaleS, a full wafer implementation, each wafer has 48 reticles with eight High-Count Analogue Neural Network (HiCANN) dice each. Each HiCANN die has 512 adaptive exponential integrate and fire (AdExp) neurons and over 100,000 synapses. The HiCANN

---

[1]Chetan Singh Thakur is with the Department of Electronic Systems Engineering, Indian Institute of Science, Bangalore, India

chip runs 10,000 times faster than real time. In SpiNNaker, ARM processors run software neural models. A 48-node SpiNNaker board is capable of simulating 250,000 neurons and 80 million synapses in real time. A recent work has successfully used the SpiNNaker system to implement a spiking neural network model of the thalamic Lateral Geniculate Nucleus (Sen-Bhattacharya et al., 2017). The IBM TrueNorth chip is designed for building large-scale neural networks and a 16-chip board is capable of running 16 million leaky-integrate-and-fire (LIF) neurons in real time. The HiAER-IFAT system has five FPGAs and four custom analogue neuromorphic integrated circuits, yielding 262k neurons and 262M synapses. The full-size HiAER-IFAT network has four boards, each of which has one IFAT module, serving 1M neurons and 1G synapses. However, all these platforms are expensive to build and require proprietary hardware and will not be easily accessible to the computational neuroscience community.

Modern FPGAs provide a large number of logic gates and physical memory, allowing large-scale neural networks to be created at a low cost. Even state-of-the-art FPGAs are affordable for research laboratories. Thus, for the past decade several projects have advanced our knowledge on how to use FPGAs to simulate neural networks. The BenNuey platform which comprises up to 18M neurons and 18M synapses was proposed by (Maguire et al., 2007). The EU SCANDLE project created a system with 1M neurons (Cassidy et al., 2011). Bailey et al. proposed a behavioural simulation and synthesis of biological neuron systems using synthesizable VHDL in (Bailey et al., 2011). The Bluehive project (Moore et al., 2012) has achieved a neural network with up-to 256k Izhikevich neurons (Izhikevich, 2003) and 256M synapses. A structured AER system, comprising 64 convolution processors, which is equivalent to a neural network with 262k neurons and almost 32M synapses, has been proposed by (Zamarreno-Ramos et al., 2012). A large-scale spiking neural network accelerator, which is capable of simulating ~400k AdExp neurons in real-time was proposed by (Cheung et al., 2016). A recent system, capable of simulating a fully connected network with up to 1440 Izhikevich neurons, was proposed in (Pani et al., 2017). We have previously presented a design framework capable of simulating 1.5 million LIF neurons in real time (Wang et al., 2014c). In that work, the FPGA emulates the biological neurons instead of performing numeric simulations as GPUs perform.

These developments strongly indicate the great potential for using FPGAs to create systems that allow research into large and complex models. Here, we present several advanced approaches based on biological observations to scale our previous system up to 2.6 billion neurons in real time. Though these models may not necessarily be at the level of detail neuroscientists would need yet, our system will present important building blocks and make important contributions in that direction. We will continue work on this system, which will eventually evolve to handle models of the complexity needed, so that our system will become a valuable neuroscience tool.

## 2. Materials and methods

### 2.1 Strategy

To emulate the neocortex (hereafter 'cortex') with billions of neurons and synapses in real time, two major problems need to be addressed: the computation problem and the communication problem. The second problem is indeed the major obstacle and far more difficult to address than the first one. Our strategy is to follow fundamental findings observed in the cortex.

### 2.1.1 Modular structure

The cortex is a structure composed of a large number of repeated units, neurons and synapses, each with several sub-types. There are six layers in the cortex; layer I is the most superficial layer and mainly contains dendritic arborisation of neurons in the underlying layers and incoming fibres from other brain areas. The distribution of neurons in the different layers is on average: 5% in layer I; 35% in layer II/III; 20% in layer IV; and 40% in layer V/VI (Dombrowski, 2001). Input to the cortex, mainly from the thalamic region, enters in layer IV; however, even in this layer, less than 10% of the afferent connections come from the thalamic region, the rest being corticocortical afferents (Martin, 2002).

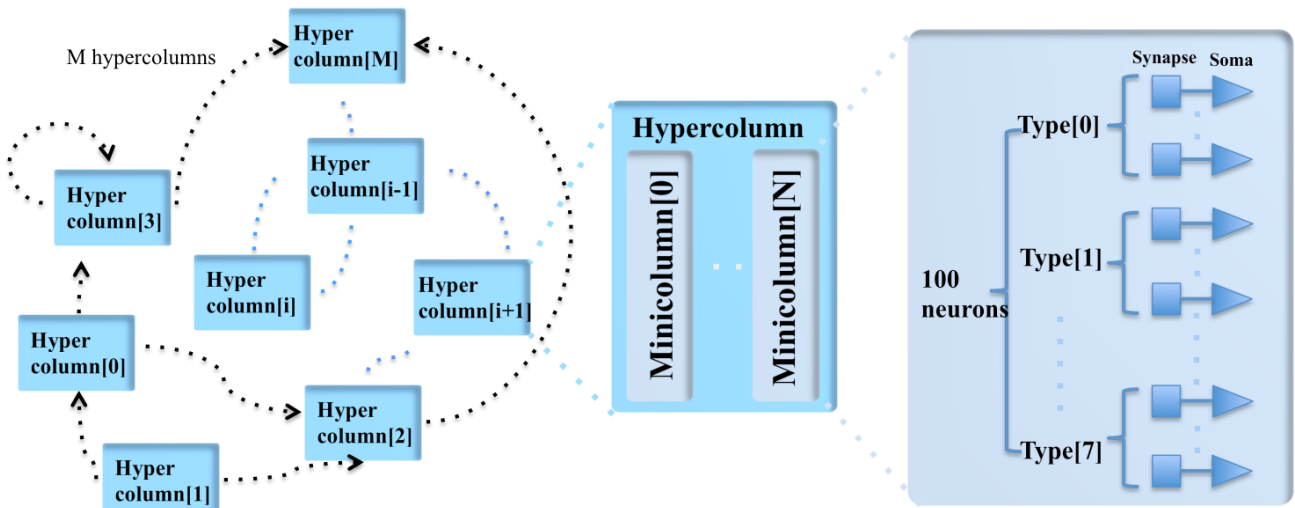

**Figure 1 | The modular structure of the cortex simulator.** The basic building block of the cortex simulator is the minicolumn, which consists of up to eight different types of heterogeneous neurons (100 in total). The functional building block is the hypercolumn, which can have up to 128 minicolumns. The connections are hierarchical: hypercolumn-level connections, minicolumn-level connections and neuron-level connections.

A minicolumn is a vertical volume of cortex with about 100 neurons that stretches through all layers of the cortex (Buxhoeveden and Casanova, 2002). Each minicolumn contains excitatory neurons, mainly pyramidal and stellate cells, inhibitory interneurons, and a large number of internal and external connections. The minicolumn is often considered to be both a functional and anatomical unit of the cortex (Buxhoeveden, 2002), and we use this minicolumn with 100 neurons as the basic building block of the cortex simulator.

As there exist some differences between minicolumns located in different parts of the cortex (such as exact size, structure, and active neurotransmitters), the minicolumn in the cortex simulator is designed to have up to eight different programmable types of neurons. The number of each type of neuron is constrained to be a multiple of four and their parameters are fully configurable while maintaining a small memory footprint. Note, the neuron types do not necessarily correspond to the cortical layers, but can be configured as such.

In the mammalian cortex, minicolumns are grouped into modules called hypercolumns (Hubel and Wiesel, 1977). These are the building blocks for complex models of various areas of the cortex (Johansson and Lansner, 2007). We therefore use the hypercolumn as a functional grouping for our simulator. Biological measurements suggest that a hypercolumn typically consists of about 100 minicolumns (Johansson and Lansner, 2007). The hypercolumn in our cortex simulator is designed to have up to 128 minicolumns. Similar to the minicolumns, the parameters of the hypercolumn are designed to be fully configurable.

### 2.1.2 Emulating dynamically

To solve the extensive computational requirement for simulating large networks, we use two approaches. First, it is not necessary to implement all neurons physically on silicon and we can use time multiplexing to leverage the high-speed of the FPGA (Cassidy et al., 2011; Wang et al., 2014a, 2014b, 2017). State-of-the-art FPGAs can easily run at a clock speed of 200 MHz (i.e., a clock period of 5 ns). Therefore, we can time-multiplex a single physical minicolumn (100 physical neurons in parallel) to simulate 200k time-multiplexed (TM) minicolumns, each one updated every millisecond. Using a pipelined architecture, the result of calculating the update for one time step for a TM neuron only has to be available just before that TM neuron's turn comes around again 1 ms later. Thus the time to compute a single TM neuron's output is not a limiting factor in the size of the network. For neural network applications that would benefit from running much faster or slower than biological neurons, we can trade off speed and multiplexing ratio (and thus network size). Limited by the hardware resources (mainly the memory), our cortex simulate was designed to be capable of simulating up to 200k TM minicolumns in real time and 1M ($2^{20}$) TM minicolumns at five times slower than real time, i.e., an update every 5 ms.

Second, based on the physiological metabolic cost of neural activity, it has been concluded that fewer than 1% of neurons are active in the brain at any moment (Lennie, 2003). Based on the experimental data in (Tsunoda et al., 2001), Johansson et al. (2007) concluded that *at most a few percent of the hypercolumns and hence only about 0.01% of the minicolumns and neurons are active in a functional sense (integrating and firing) at any moment in the cortex.* Hence, in principle, one hardware minicolumn could be dynamically reassigned to simulate $10^4$ minicolumns on average. Such a hardware minicolumn will be referred to as a physical minicolumn and the minicolumn to be simulated will be referred to as a dynamically assigned (DA) minicolumn. If a DA minicolumn cannot be simulated in a single time step, the physical minicolumn needs to be assigned to that DA minicolumn for a longer time and the number of DA minicolumns that can be simulated will decrease proportionally.

Theoretically, a TM minicolumn array with 1M TM minicolumns can be dynamically assigned for $1M \times 10^4 = 10^{10}$ minicolumns, if these minicolumns can be updated every 5 ms and if only 0.01% of the minicolumns are active at any time step. To be able to deal with much higher activity rates, we chose to dynamically assign one TM minicolumn for 128 DA minicolumns. The maximum active rate of the minicolumns that this system can support is therefore $1/128 \approx 0.7\%$, allowing it to support $128 \times 1M = 128M$ ($2^{27}$) minicolumns (each has a unique 27-bit address). This means the simulator is capable of simulating up to 1M hypercolumns, each of which has up to 128 minicolumns. We can trade off the active rate and the network size when needed. This dynamic-assignment approach has been validated in our previous work (Wang et al., 2013b, 2014d, 2015). Applying this approach with nanotechnology has been fully explored in (Zaveri and Hammerstrom, 2011).

### 2.1.3 Hierarchical communication

Our cortex simulator uses a hierarchical communication scheme such that the communication cost between the neurons can be reduced by orders of magnitude. It is impractical to use point-to-point connections for cortex-level simulation, as it would require hundreds of terabytes of memory to specify the connections and their weights. Anatomical studies of the cortex presented in (Thomson, 2003) showed that cortical neurons are not randomly wired together and that the connections are quite structural. Connections of the minicolumns are highly localised so that the connectivity between two pyramidal neurons less than 25–50 μm apart is high and the connectivity between two neurons drops sharply with distance (Holmgren et al., 2003). Furthermore, pyramidal neurons in layer II/III and V/VI within a minicolumn are reciprocally connected (Thomson, 2003).

Based on these findings, we chose to store the connection types of the neurons, the minicolumns, and the hypercolumns in a hierarchical fashion instead of individual point-to-point connections. In this scheme, the addresses of the events consist of hypercolumn addresses and minicolumn addresses. Both of them are generated on the fly with connection parameters according to their connection levels respectively. For example, we can specify that hypercolumns 1 to 10000 are each connected with random weights to the 10 hypercolumns closest to them, while hypercolumns 20000 to 40000 are each connected their nearest 16 hypercolumns with fixed weights. Proximity is defined by the distance of the hypercolumn index in address space. This method only requires several kilobytes of memory and can be easily implemented with on-chip SRAMs. We will present the details of this hierarchical scheme in section 2.6.

Inspired by observations from neurobiology, where neurons and their connections often form clusters to create local cortical microcircuits (Bosking et al., 1997), the communication between the neurons uses events (spike counts) instead of individual spikes. This arrangement models a cluster of synapses formed by an axon onto the dendritic branches of nearby neurons. The neurons of one type within a minicolumn all receive the same events, which are the numbers of the spikes generated by one type of neuron in the source minicolumns within a time step.

One minicolumn has up to eight types of neurons, and each type can be connected to any type of neuron in the destination minicolumns. Hence, there are up to 64 synaptic connections possible between any two minicolumns. We impose that a source minicolumn has the same number of connections to all of the other minicolumns within the same hypercolumn, but that these can have different synaptic weights. The primary advantage of using this scheme is that it overcomes a key communication bottleneck that limits scalability for large-scale spiking neural network simulations. Otherwise, we would be required to replicate spikes for each group of target neurons.

Our system allows the events generated by one minicolumn to be propagated to up to 16 hypercolumns, each of which has up to 128 minicolumns, i.e., to 16×128×100 = 200k neurons. Each of these 16 connections has a configurable fixed axonal delay (from 1 ms to 16 ms, with a 1 ms step). Since a hypercolumn has up to 128 minicolumns, a hypercolumn can communicate with far more than 16 hypercolumns. Further details will be presented in section 2.5.

There two major differences between our scheme and the hierarchical address-event routing scheme used by the HiAER-IFAT system. First, HiAER routes each individual spike while we route the events (the numbers of the spikes generated by one type of neuron (in a minicolumn). Second, HiAER uses external memory to store its look up tables (LUTs) such that it is capable of supporting point-to-point connections, each of which has its own programmable parameters, e.g., synaptic weight and axonal delay. This allows HiAER-IFAT to be a general purpose neuromorphic platform for spike-based algorithms. Its scalability is therefore constrained by the size of the external memories. While our scheme takes the advantage of the structural connections of the cortex, we use the on-chip SRAMs to implement LUTs that only store the connection patterns.

## 2.2 Architecture

The cortex simulator was deliberately designed to be scalable and flexible, such that the same architecture could be implemented either on a standalone FPGA board or on multiple parallel FPGA boards. As a proof-of-concept, we have implemented this architecture on a Terasic DE5 kit (with one Altera Stratix V FPGA, two DDR3 memories and four QDRII memories) as a standalone system. Figure 2 shows its architecture, consisting of a neural engine, a Master, off-chip memories, and a serial interface.

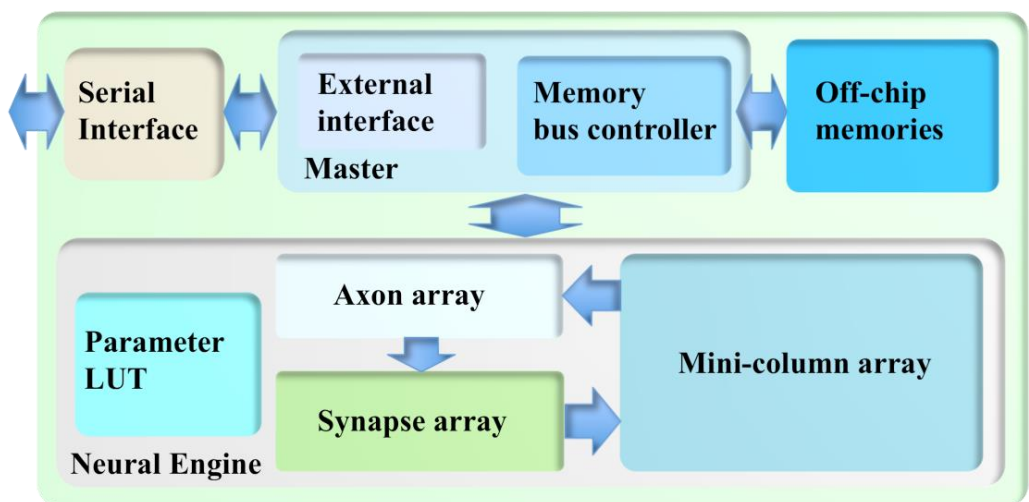

**Figure 2** | **The architecture of the cortex simulator.** The system consists of a neural engine, a Master, off-chip memories and a serial interface. The neural engine realises the function of biological neural systems by emulating their structures. The Master controls the communication between the neural engine and the off-chip memories, which store the neural states and the events. The serial interface is used to interact with the other FPGAs and the host controller, e.g., PCs.

The neural engine forms the main body of the system. It contains three functional modules: a minicolumn array, a synapse array, and an axon array. The minicolumn array implements TM minicolumns. The axon array will propagate the events generated by the minicolumns with axonal delays to the synapse array. In the synapse array, these events will be modulated with synaptic weights and will be assigned their destination minicolumn address. The synapse array will send these events to the destination minicolumn array in an event-driven fashion. Besides these functional modules, there is a parameter look-up table (LUT), which stores the neuron parameters, connection types, and connection parameters. We will present the details of these modules in the following sections.

Because of the complexity of the system and large number of the events, each module in the neural engine was designed to be a slave module, such that a single Master has full control of the emulation progress. The Master has a memory bus controller that will control the access of the external memories. As we use time multiplexing to implement the minicolumn array, we will have to store the neural state variables of each TM neuron (such as their membrane potentials). These are too big to be stored in on-chip memory and have to be stored in off-chip memory, such as the DDR memory. Using off-chip memory needs flow control for the memory interface, which makes the architecture of the system significantly more complex, especially if there are multiple off-chip memories. The axon array also needs to access the off-chip memories for storing events. We will present the details of the Memory Bus Controller in section 2.7.1.

The Master also has an external interface module that will perform flow control for external input and output. This module also takes care of instruction decoding. We will present the details of External Interface in section 2.7.2. The serial interface is a high-speed interface, such as the PCIe interface, that communicates with the other FPGAs and the host PC. It is board-dependent, and in the work presented here, we use Altera's 10G base PHY IP.

## 2.3 Minicolumn array

The minicolumn array consists of a neuron-type manager, an event generator, and the TM minicolumns, which are implemented by time-multiplexing a single physical minicolumn consisting of 100 physical neurons. These neurons will generate positive (excitatory) and negative (inhibitory) post-synaptic currents (EPSCs and IPSCs) from input events weighted in the synapse array. These PSCs are integrated in the cell body (the soma). The soma performs a leaky integration of the PSCs to calculate the membrane potential and generates an output spike (post-synaptic spike) when the membrane potential passes a threshold, after which the membrane potential is reset and enters a refractory period. Events, i.e., spike counts, are sent to the axon array together with the addresses of the originating minicolumns, the number of connections, and axonal delay values for each connection (between two minicolumns).

### 2.3.1 Physical neuron

Rather than using a mathematical computational model with floating-point numbers, the physical neuron has been efficiently implemented using fixed-point numbers to minimise the number of bits that need to be stored for each state variable. The minimisation of memory use was effectively achieved using a stochastic method to implement exponential decays for the conductance-based neuron model (Wang et al., 2016), (Wang et al., 2014c). We store only the most significant bits (MSBs) and generate the least significant bits (LSBs) on the fly with a random number generator. This method not only reduces the storage needed, but also introduces variability between the (TM) neurons while using the exact same physical neuron model. This makes the network simulations more realistic. In existing implementations, digital neurons are usually perfectly matched, so that two neurons in the network with identical initial conditions receiving the same input will remain synchronised forever.

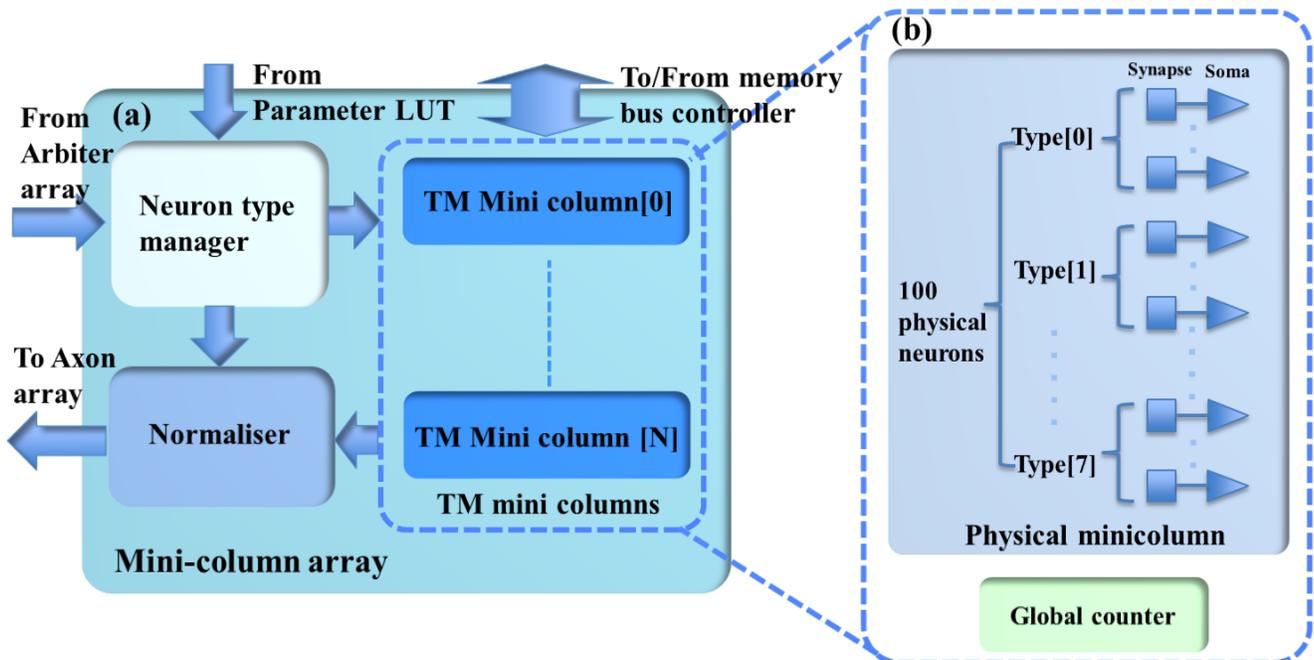

**Figure 3 | (a) The structure of the minicolumn array. (b) The internal structure of the TM minicolumns.** The TM minicolumns are implemented with the time-multiplexing approach and Figure 3b shows its internal structure. It consists of a physical minicolumn and a global counter. The global counter processes the TM minicolumns sequentially. The physical minicolumns consists of 100 physical neurons.

The stochastic conductance-based neuron is a compact design using only fixed-point numbers. It has a single post-synaptic current generator (see Figure 4), which can generate both EPSCs and IPSCs, and a soma to integrate the post-synaptic currents. The PSC generator consists of two multipliers, an adder, a comparator and two multiplexers. Its function is expressed by the following equation:

$$PSC(t+1) = PSC(t) \times \frac{\tau_{psc}}{\tau_{psc}+1} + r(t) + g_{syn} \times W(t) \quad (1)$$

where $t+1$ represents the index of the current time step, $PSC(t)$ represents the value of the PSC (from the memory), and $W(t)$ represents the linear accumulation of the weighted synaptic input from the pre-synaptic events within previous time step. Both $PSC(t)$ and $W(t)$ are signed 4-bit numbers and normalised in the range [-1,1]. $\tau_{PSC}$ represents the time constant, i.e., 10 ms, controlling the speed with which $PSC$ will decay to 0 exponentially. The PSC generator will use $\tau_{EPSC}$ and $\tau_{IPSC}$ for EPSC and IPSC respectively. $r[t]$ is a random number drawn from a uniform distribution in the range (0,1), and it will be different at different time steps, even for the same TM neuron. $g_{syn}$ is the synaptic gain, and is an 8-bit number normalised in the range [0.06,16].

In the hardware implementation, the time constants are efficiently implemented by applying a simple shift operation to the product of $PSC(t)$ and a leakage rate $L$ (an 8-bit number, $L/256 \approx \tau/\tau+1$). $r$ is a 5-bit random number and the maximum time constant it can achieve is 30 ms (30 time steps). Its implementation can easily be modified if longer time constants are needed.

The soma is also a stochastic conductance-based model similar to the PSC generator. Its function is expressed similarly by the following equation:

$$V_{mem}(t+1) = V_{mem}(t) \times \frac{\tau_{soma}}{\tau_{soma}+1} + r(t) + g_{psc} \times PSC(t+1) \quad (2)$$

where $V_{mem}(t)$ represents the previous value of the membrane voltage (from the memory). The soma has two states: active state and the refractory state. The soma can only integrate $PSC(t+1)$ in its active state: its membrane voltage is greater than its initial value (a configurable parameter). Otherwise, the soma is in its refractory state and the PSC will be discarded. $\tau_{soma}$ represents the time constant controlling the speed with which $V_{mem}$ will decay to the initial value exponentially. The soma will use $\tau_{mem}$ and $\tau_{rfc}$ for the active and the refractory period, respectively. $r[t]$ is also a random number drawn from a uniform distribution in the range (0,1). $g_{psc}$ is the synaptic gain, and it is an 8-bit number normalised in the range [0.06,16]. The soma will generate a post-synaptic spike when $V_{mem}$ overflows and $PSC(t+1)$ is an EPSC. $V_{mem}$ will be reset, when there is an overflow/underflow caused by EPSC or IPSC respectively.

The physical neuron has an 11-stage pipeline without halt. Each TM neuron will access each computing module such as the adder and the multipliers for only one clock cycle. In one clock cycle, the adder and the multipliers are all being used, but by different TM neurons. The overhead of the pipeline is negligible, i.e., 22 clock cycles: the first 11 cycles for setting up the pipeline and the last 11 cycles for waiting for the last TM neuron to finish computing.

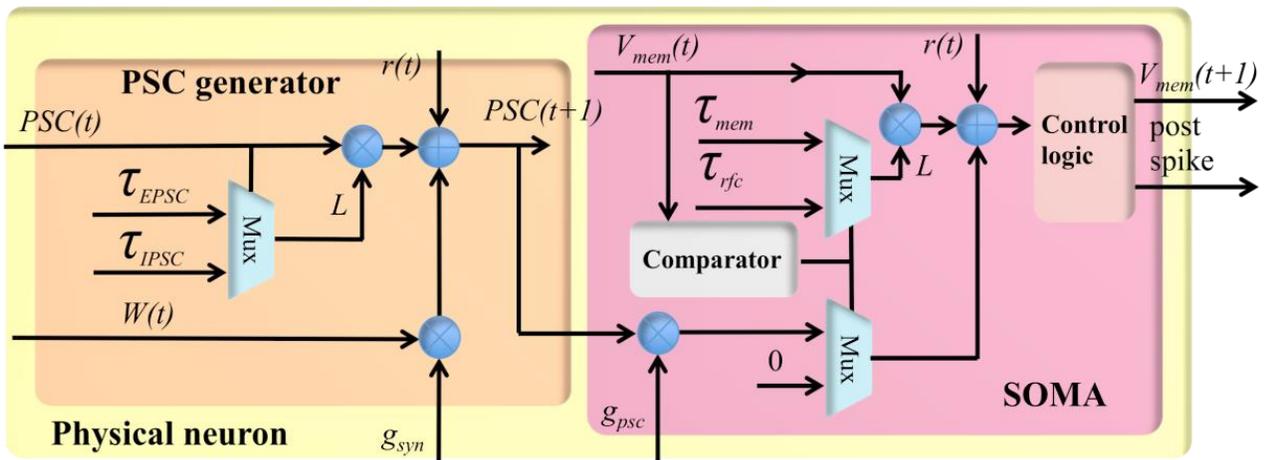

**Figure 4. The structure of the physical neuron.**

### 2.3.2 Time multiplexing

Time multiplexing enables implementing large-scale spiking neural networks, while requiring only a few physical neurons, by leveraging the high-speed of the FPGA. The bottleneck for time multiplexing is data storage. The on-chip memory is typically highly limited in size (generally only several megabytes). In our system, one physical neuron requires 8 bits and one TM minicolumn thus requires 800 bits. Hence, a network with 100M TM minicolumns would require 800 Mb, which is too big for on-chip memory. We therefore need to use off-chip memory to store these variables. The use of off-chip memory, however, will be limited by the communication bandwidth and will need flow control for the memory interface, which will make the architecture of the system more complex.

We have developed a time-multiplexing approach that is capable of using off-chip memory efficiently. Here, the whole TM system is segmented into blocks of 1024 (1k) TM neurons. A ping-pong buffer is used in the memory interface such that the system will run one segment at a time, and then pause until the new values from the off-chip memory are ready for the next segment. This allows us to use the external DDR memory in burst mode, which has maximum bandwidth. Thus, the size of a segment was chosen to be equal to the burst size of the DDR memory. Further details about the memory interface will be presented in section 2.7.1.

### 2.3.3 Neuron-type manager

As each minicolumn can have up to eight different types of neurons, a neuron-type manager is introduced to assign the parameters and weights to each individual neuron. The neuron parameters, which are stored in the parameter LUT, include following parts:

1. The number of types of neurons in the minicolumn
2. The number of each type of neuron
3. The parameters of each type of neuron

These functions are efficiently implemented with an 8-to-25 multiplexer (25 = 100/4, each minicolumn has 100 neurons and the number of each type of neuron is constrained to a multiple of 4). The neuron-type manager will use linear feedback shift registers (LFSRs) to generate random numbers ($r(t)$) for each physical neuron such that the neurons within a TM minicolumn, even those with identical parameters, are heterogeneous.

### 2.3.4 Events generator

For each type of neuron within a TM minicolumn, the spikes generated by the physical neurons are summed to produce spike counts. Since the source neuron type and the destination neuron types could have different numbers of neurons, these spike counts will be limited to 4-bit numbers. This summation is implemented with eight parallel adders, each of which has 25 inputs, in a 3-stage pipeline. The number of each neuron type is obtained from the neuron-type manager via shift registers, which were designed to work with the pipelines of the physical neuron.

## 2.4 Axon array

The axon array propagates the events from the minicolumn array or from the external interface to the synapse array, using programmable axonal delays. To implement this function on hardware, we used a two-phase scheme comprising a TX-phase (TX for transmit) and a RX-phase (RX for receive). In the TX-phase, the events are written into different regions of the DDR memories according to their programmable axonal delay values. In the RX-phase, for each desired axonal delay value, the events are read out from the corresponding region of the DDR memories stochastically, such that their expected delay values are approximately equal to the desired ones.

The DDR memories are all single-port devices; hence, this two-phase scheme will access them in an alternating fashion such that these two phases do not try to access the same DDR device at the same time. This is guaranteed because the TX-phase has higher priority for accessing the DDR memories: the TX-phase will choose one of the DDR memories to write the events first, and then the RX-phase will choose one of the remaining DDR memories to read. This implementation was chosen because, if the TX-phase cannot move the events generated by the minicolumn array quickly enough, the Master will have to pause the minicolumn array to avoid an overflow of events.

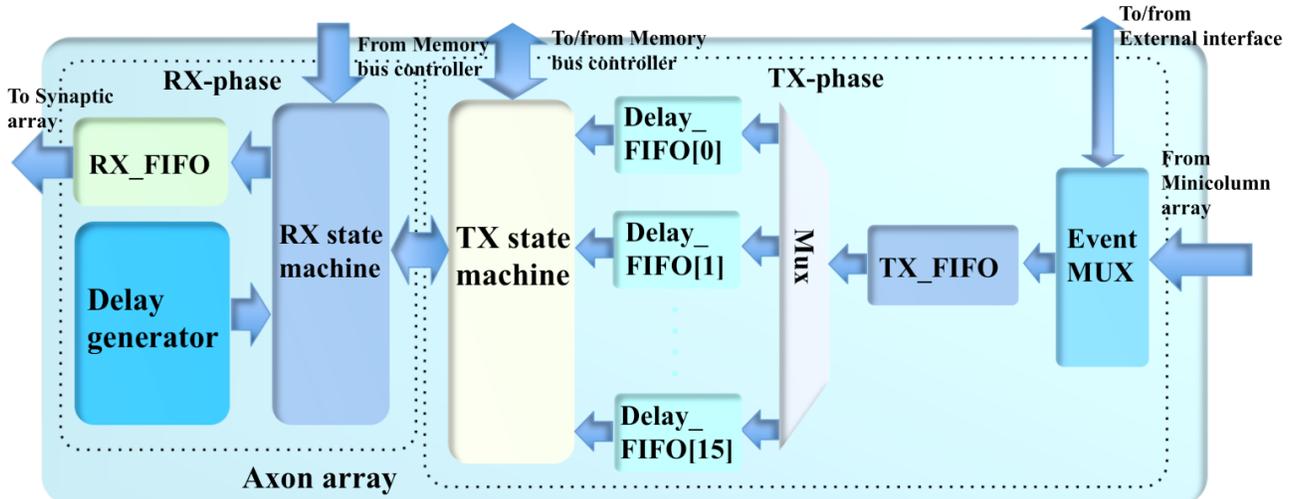

**Figure 5. The structure of the axon array.**

In the system presented here we have two identical 8 GB DDR memories: DDR_A and DDR_B, each of which has eight regions (~1 GB each, one for each axonal delay value). DDR_A is for the eight odd-number axonal delay values, e.g., 1 ms, 3 ms, and so on. DDR_B is for the eight even-number axonal delay values. This arrangement is designed to match the alternating access of the DDR memories. Note, the memory bus controller will only grant the axon array access to the DDR memories when the minicolumn array is not accessing them. Further details of this arrangement will be presented in section 2.7.1.

### 2.4.1 TX-phase

The TX-phase consists of an event multiplexer, a TX_FIFO (first in first out), a TX state machine and 16 Delay_FIFOs (one per axonal delay value). The event multiplexer transfers the events from the minicolumn array and the external interface to the TX_FIFO. The event multiplexer also sends the events from the minicolumn array to the external interface in case these events need to be sent to other boards (for a multi-FPGA system). The internal events (from the minicolumn array) will be directly written to the TX_FIFO. The event multiplexer simply needs to hold the external events when there are internal events.

The events from the TX_FIFO are moved to the Delay_FIFOs according to their axonal delay values. To avoid losing events, the Master will pause the minicolumn array when the TX_FIFO is almost full. The Delay_FIFO with the maximum usage will be selected by the TX state machine for the next write operation. This selection will decide the target DDR memory automatically. The TX state machine will try to write all the events in the selected Delay_FIFO to the DDR memory as long as the size does not exceed the allowed burst length.

To completely eliminate any potential collisions with the RX-phase, the TX and RX state machines use a cross-lock handshake scheme such that TX-phase will only be allowed to initiate choosing the Delay_FIFO and start writing the selected DDR memory when the RX-phase is inactive (not reading). Otherwise, the TX-phase will wait for the RX-phase to finish reading. The RX-phase will not initiate until the TX-phase has started, unless all the Delay_FIFOs are empty. The TX state machine will notify the memory bus controller of the completion of one TX-phase. This procedure will continue to repeat as long as the axon array is allowed to access the memory.

### 2.4.2 RX-phase

The RX-phase consists of an RX state machine, an RX_FIFO, and a delay generator. The delay generator will enable the RX state machine to stochastically read the events from the DDR memory at different rates to achieve the desired axonal delay values approximately. For each axonal delay value, the events are read out with a probability $P_i$, which meets the following conditions:

$$\sum_{i=1}^{16} P_i = 1 \qquad (3)$$

$$P_i = P_1/i \qquad (4)$$

Equation (4) ensures that the ratios between the mean times for reading the same number of events match the ratios between the axonal delay values. For instance, if it takes $T$ time to read out $N$ events with the axonal delay value of 1 ms, it will take ~$16T$ time to read the same number of events with an axonal delay value of 16 ms. This assumption requires two conditions to be true: first, the DDR memories have been filled with sufficient events; second, the read-out operations must be evenly distributed across all the 16 axonal delay values. The first assumption is always true as long as the minicolumn array is generating events. To meet the second condition, all the probabilities $P_i$ are normalised to 20-bit integers $R_i$, which are used to generate a set of subsequent thresholds $T_i$ (17 20-bit integers and $T_0 = 0$) such that $T_{i+1} - T_i = R_i$. In hardware, in each clock cycle, a 20-bit LFSR generates a random value that will fit in one of the threshold ranges (between $T_{i+1}$ and $T_i$). This implies that in each clock cycle, the delay generator will select one of the 16 axonal delay values (of which the events should be read out). If the selected axonal delay value has no events to be read out, we simply wait for the next clock cycle. With no complicated rules involved, this implementation effectively distributes the read-out operation evenly across all the axonal delay values. Note that using 20-bit integers is an arbitrary choice. In theory, the higher the resolution, the more even the distribution will be; however, this requires more logic gates, and our result shows that a 20-bit resolution is sufficient.

The above approach only provides the ratios of delays; to achieve the actual delay values, we need to scale $P_i$ with a reconfigurable global probability $f$. This function is implemented with a stochastic approach that does not involve multiplication. The read out is only enabled when the 10-bit normalised $f$ is no greater than the value of a 10-bit LFSR. In hardware, the axon array will access the DDR memories every 1k/200 MHz≈5 μs. If $f$ is set to 0, the mean of the actual '1 ms' axonal delay will be ~5 μs. If $f$ is set to 1023 (0x3FF), the mean of the actual 1 ms axonal delay will be 5 μs×1023≈5 ms. The value of $f$ also can be adjusted for the firing rate of the minicolumns for more accurate emulation.

As the TX-phase will select one DDR memory, the RX state machine will initiate a read operation immediately after the start of the TX-phase by selecting the axonal delay values in the other DDR memory. If there is no event in the region selected by the delay generator, we simply wait for the next selection in the next clock cycle. The events read from the DDR memories are written into the RX_FIFO, which are read out by the synapse array. The RX-phase will not be initiated if the RX_FIFO is full. The data flow in the RX-phase is far less critical than that in the TX-phase since the capacities of the DDR memories are orders of magnitude larger than the on-chip memory.

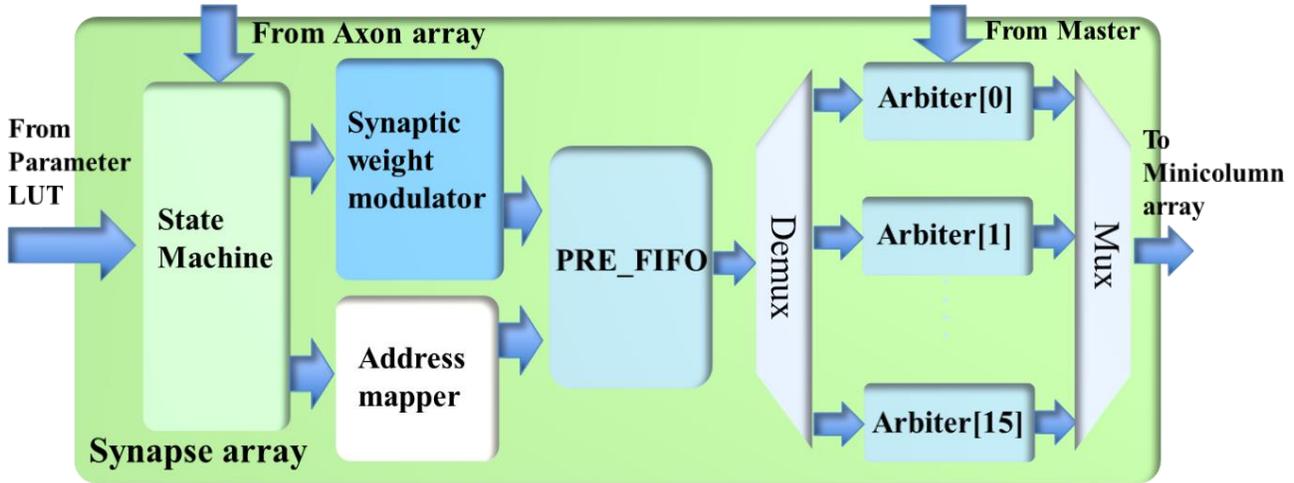

**Figure 6** | **The structure of the synapse array.**

## 2.5 Synapse array

The synapse array (see Figure 6) emulates the function of biological synaptic connections: it modulates the incoming events from the axon array with synaptic weights and generates destination minicolumn addresses for them. These events are then sent to the TM minicolumns. The synapse array only performs the linear accumulation of synaptic weights of the incoming events, whereas the exponential decay is emulated by the PSC generator in the neuron.

### 2.5.1 Processing events

The synapse array consists of a state machine, a synaptic weight modulator, an address mapper, a PRE_FIFO (the FIFO for storing pre-synaptic events), and 16 arbiters. The axon array has already performed the function of converting one post-synaptic event to multiple pre-synaptic events according to its connections. Hence, one incoming pre-synaptic event to this synapse array will be sent to the minicolumns of one and only one hypercolumn, which are referred to as the destination minicolumns and the destination hypercolumn, respectively. For instance, the destination hypercolumn has 100 minicolumns and the size of the connection is 32 (minicolumns); thus, 32 out of 100 minicolumns will be randomly selected as the destination minicolumns.

A serial implementation that performs this function would take up to 128 cycles and would be extremely slow. A parallel implementation that performs this function in one cycle would be too big and unnecessary, since not all connections have 128 minicolumns. As a compromise, this function is performed in a time slot of four clock cycles: the synaptic weight modulator and the address mapper process up to 32 connections in each clock cycle using a 12-stage pipeline. For instance, if the size of the connections is 80, we will process 32, 32, 16 and 0 connections in the four time slots, respectively. Although the number of connections could be less than 128, the time slot has to be fixed to four clock cycles as pipelines can only achieve the best performance with fixed steps, else the overhead of pipeline will be significant.

The state machine reads one pre-synaptic event from the axon array if the RX_FIFO of the axon array is not empty. The minicolumn address of the incoming event is sent to the parameter LUT for its connection parameters. With these parameters, the address mapper first generates the destination hypercolumn address (20 bits) by adding an offset (20 bits) to the hypercolumn address (20 bits) of the incoming event. If it overflows, it will wrap. Recurrent connections can be realised by setting the offset to zero. The addresses of the destination minicolumns are generated randomly based on the size of the connections. The events with a given address are sent to the same destination minicolumns at each time.

Along with the address mapper, the synaptic weight modulator first multiplies the eight spike counts (of the 8 neuron types) of this event with the eight corresponding synaptic weights. To update the 64 synaptic connections between two minicolumns, for each type, these eight weighted spike counts are linearly summed with an 8-bit mask (1 bit per connection). The results of this operation are written into the PRE_FIFO, which is read out by the arbiters.

### 2.5.2 Arbiter

The arbiters read the events from the PRE_FIFO and assign them to the TM minicolumns. To successfully perform this task, it needs to figure out which TM minicolumn to assign (for one incoming event) within a short time, e.g., tens of clock cycles. If not, the FIFO of the arbiter that holds the newly arrived events during processing will soon be full and alert the Master to stop the synapse array from reading events.

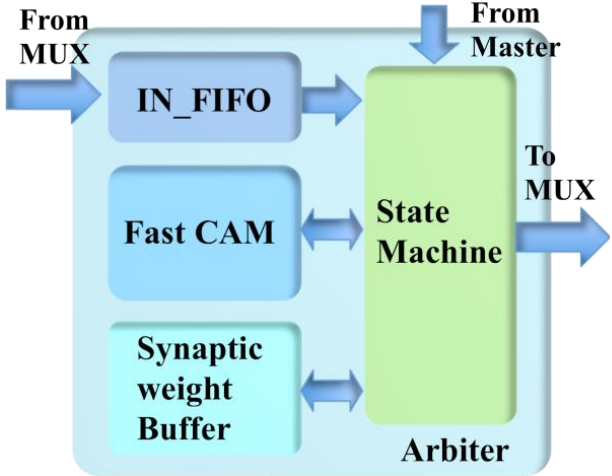

**Figure 7 | The structure of the arbiter.**

In principle, one single arbiter for all the TM minicolumns can achieve the best utilisation as any TM minicolumn can be assigned to any incoming event. A straightforward implementation for this approach was to use an address register array (Wang et al., 2013b, 2014d), which is effectively a content-addressable memory (CAM). It is capable of checking whether there already is a TM minicolumn assigned to the incoming event or not in each clock cycle. However, this approach uses a large number of flip-flops. Since the limited number of flip-flops is one of the bottlenecks for large-scale FPGA designs, this simple method is not possible for 1M TM minicolumns.

To increase parallelism, 16 identical arbiters are used such that each of them can perform the dynamic assignment function for 1M/16 = 65536 (64k) TM minicolumns. The events in the PRE_FIFO are moved to these arbiters according to the most significant four bits of their addresses. The potential loss in utilisation of the TM minicolumns will be minimal as long as the (DA) minicolumns are evenly distributed across the whole address range. This requirement is not difficult to meet for large-scale and sparsely connected neural networks with low activity rates.

The arbiter consists of a state machine, a fast CAM, a synaptic weight buffer, an address buffer, and an IN_FIFO, which is introduced to avoid losing spikes that arrive during the processing time. The fast CAM (assuming its size is 64k×23 bit) stores the addresses of minicolumns that have been assigned to the TM minicolumns. As a CAM, it compares input search data against stored data, and returns the address of matching data, which is then used to access the synaptic weight buffer. The latter is a dual-port SRAM with a size of 64k×32 bit (one 32-bit value per TM minicolumn).

The state machine reads the IN_FIFO when it is not empty and processes its output before another read. Inside the state machine, there is an index generator implemented by a counter that controls the fast CAM and the synaptic weight buffer. The output value of this index generator is used as the address of the TM minicolumns. The state machine accesses the fast CAM to check whether there is already a TM minicolumn assigned to this event or not. If there is already a TM minicolumn assigned to the incoming event, its values are linearly accumulated to the values stored in the synaptic weight buffer. Otherwise, the state machine sends this event to an unassigned TM minicolumn, as indicated by the current value of the index generator, by storing the address and weights of this event in the fast CAM and the synaptic weight buffer, respectively. This TM minicolumn is then labelled as assigned by the state machine. After finishing the above actions, the index generator is incremented by one. Its count is reset to zero after it reaches the maximum value.

The most challenging part is the implementation of the fast CAM, which has to complete its search within tens of clock cycles. A straightforward implementation using 64k 23-bit flip-flops, one per TM minicolumn, would be capable of performing this task. However, the amount of flip-flops required would too large for the FPGA. An implementation using a single SRAM would be too slow since it can access only one value per clock cycle. In the worst case, the search time would be 64k clock cycles.

To overcome this problem, we used multiple SRAMs to implement the fast CAM. Because hypercolumns have many minicolumns, we use each value of the fast CAM for eight TM minicolumns simultaneously. In this way, we can reduce the size of the fast CAM from 64k×23 bits to 8k×20 bits. It can be implemented efficiently using 128 single-port SRAMs, each with a size of 64×20 bits. The maximum search time is thus reduced to 64 clock cycles. The 16-bit address for accessing the synaptic weight buffer is obtained by concatenating the 13-bit address returned by the fast CAM with the least significant 3 bits of the address of the event.

Because the events for the destination minicolumns of the same hypercolumn arrive sequentially, we introduce a bypass mechanism that is capable of significantly reducing the processing time. The newly-read event from the IN_FIFO is first compared with the previous event, before it is sent to the fast CAM. If the most significant 20 bits of their addresses match, the state machine bypasses the fast CAM and performs linear accumulation directly. The values stored in the synaptic weight buffer are accumulated until they are read out and cleared by the minicolumn array.

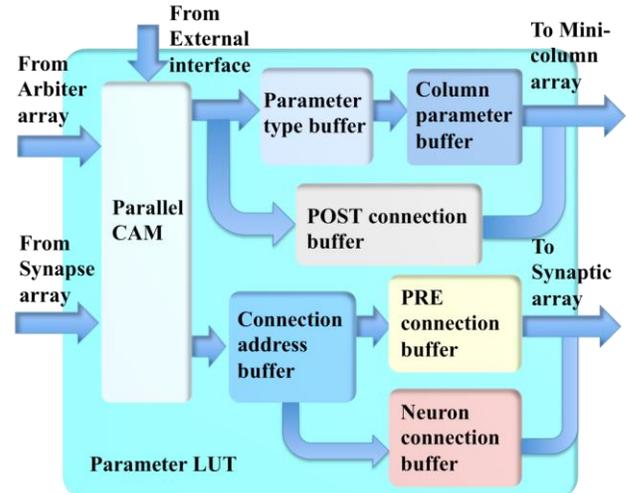

**Figure 8. The structure of the parameter LUT.**

To obtain the parameters of the neurons and hypercolumn connections, the minicolumn array also needs to read the values stored in the fast CAM, which are sent to the parameter LUT. The read-out of the CAM uses the same scheme as the one for the synaptic weight buffer but without the 'clear' operation.

## 2.6 Parameter LUT

The parameter LUT performs the look-up table function for the incoming addresses from the synapse array and minicolumn array within several clock cycles. Its key component is a parallel CAM, which is implemented with 512 27-bit flip-flops, instead of with SRAM, such that it can be accessed by more than one input simultaneously with a fixed latency of only three clock cycles. This fixed latency requirement is critical and must be met, because the system is a highly pipelined design, which will not work with non-fixed latencies. The latency of an SRAM-based CAM is generally not fixed.

The conventional CAMs compare input search data against stored data and return the address of matching data. The parallel CAM differs slightly from the conventional CAMs: it stores a set of sequential thresholds $A_i$ and its return value tells in which range (between $A_{i+1}$ and $A_i$) the input address is. This return value is used for accessing the buffers.

The minicolumn array requires two sets of parameters. The first set is for computing the neuron type, time constants and gains of the neurons. Two buffers are used for reducing memory usage. The returned address from the parallel CAM is used to access the parameter-type buffer, the values stored in it are then used to access the minicolumn parameter buffer. The second set of parameters is for routing the post-synaptic events: the axonal delay values and the hypercolumn connections, which are stored in the post-connection buffer.

For the synapse array, the returned addresses from the parallel CAM are used to access the connection address buffer. The values stored in this buffer are then used to access two other buffers. The first buffer is a pre-connection buffer, which stores the size of the connections, offset, synaptic weights, and the size of the destination hypercolumn. The second buffer is a neuron-connection buffer, which stores the synaptic connections between two minicolumns.

## 2.7 Master

The Master plays a vital role in the cortex simulator: it has complete control over all the modules in the neural engine such that it can manage the progress of the simulation. This mechanism effectively guarantees no event loss or deadlock during the simulation. The Master slows down the simulation by pausing the modules that are running quicker than other modules. The Master has two components (see Figure 2): a memory bus controller and an external interface.

### 2.7.1 Memory bus controller

The memory bus controller has two functions: interfacing the off-chip memory with Altera IPs, and managing the memory bus sharing between the minicolumn array and the axon array. There are multiple off-chip memories in the system: two DDR memories (128M×512 bits with a 512-bit interface) and four QDRII memories, each of which is a high-speed dual-port SRAM but with a limited capacity (1M×72 bits, with a 72-bit interface). Storage of the neural states of the minicolumn array will use both the DDR and QDRII memories. For the DDR memories, the lowest 512k×512 bits are used for neural states. The rest of the DDR memories (~8 GB) are used to store the post-synaptic events (evenly divided by eight, one per delay), as presented in section 2.4.

Accessing the QDR memories is quite straightforward as they are all dual-port devices. However, they are running in a different clock domain than the DDR memories, so that we need two asynchronous FIFOs (QDR WR_FIFO and RD_FIFO) for communication between the two clock domains. To efficiently access the DDR memories for the neural states, the memory bus controller accesses them alternately such that while writing 1k neural states to one DDR memory, 1k neural states from the other one are read. In the next burst operation, these roles are reversed. This method effectively uses two single-port memories to build a dual-port memory with a bandwidth of 512 bits@200MHz, whereas four QDR memories can provide 288 bits@225MHz. Hence, we can provide enough bandwidth (800 bits@200MHz) for the TM minicolumns (100 physical neurons, each neuron needs 8 bits).

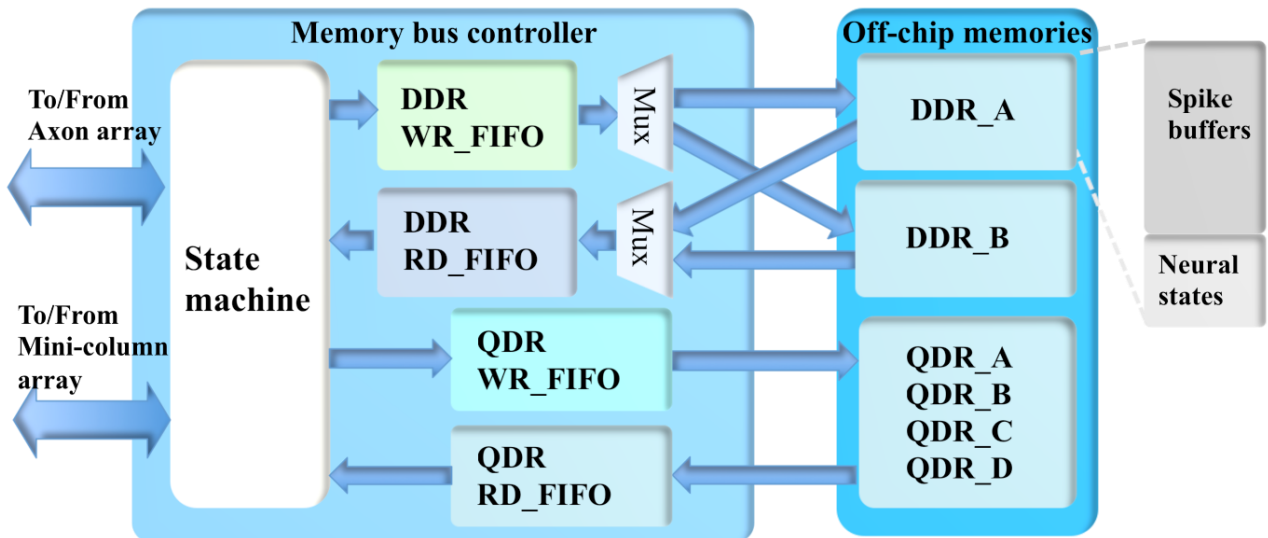

**Figure 9. The structure of the memory bus controller.**

There are waiting times when reading the DDR memories. The TM system is segmented using a size of 1024 (1k) TM minicolumns, and there is no waiting time within one segment. To make sure the neural states from the memories are always available during one segment, we use a read-in-advance scheme: the neural states being read out in the current segment are to be used in the next segment. These 1k neural states will be written into a DDR RD_FIFO, which sends its output to the minicolumn for computing. The write operation is relatively less critical because we can choose when to write to the DDR memory. We simply write the neural states from the minicolumn array into the DDR memory via a DDR WR_FIFO. It plays the role of a buffer, since there are waiting times when writing the DDR memory.

The memory bus controller will only grant the axon array access to the DDR memory during the slots between two 1k-burst operations. In the system presented here, this slot is configured to be up to 200 clock cycles. This means if the axon array finishes its operation within 200 clock cycles, it will release this access after that. To avoid the loss of events, when the usage of TX_FIFO (of the axon array) exceeds 95%, the memory bus controller will pause the minicolumn array by not allowing it access to the off-chip memory. Instead, the memory bus controller will allow the axon array to keep using the bus to write events to the DDR memory until the usage of the TX_FIFO is lower than 95%.

**TABLE I. Device utilisation on Altera Stratix 5SGXXEA7N2F45C2**

| Adaptive Logic Modules (ALMs) | RAMs | DSPs |
|---|---|---|
| 157978/234720 | 41 Mbit/52 Mbit | 256/256* |

*One DSP slice can be used to implement three 9-bit fixed-point multipliers. The Altera synthesise tool will only use this function when all the DSPs have been used.

### 2.7.2 External interface

The external interface module will control the flow of the input and output of events and parameters. The main input to this system are events, which are sent to the minicolumn array via the axon array. This module also performs instruction decoding such that we can configure the parameter LUT and the system registers.

The outputs of the simulator are individual spikes (100 bits, one per neuron) and events generated by the minicolumns. The individual spikes (with their minicolumn addresses) are sent to the host via an SPK_FIFO if their addresses are labelled as being monitored. To support multi-FPGA system, the events that labelled to other boards will also be sent to the host.

## 3. Results

### 3.1 Prototype system

The system was developed using the standard ASIC design flow, and can thus be easily implemented with state-of-the-art manufacturing technologies, should an integrated circuit implementation be desired. A bottom-up design flow was adopted, in which we designed and verified each module separately. Once the module-level verification was complete, all the modules were integrated together for chip-level verification. Table I shows the utilisation of hardware resources on the FPGA. Note that this utilisation table includes the Altera IPs such as the ones for accessing external memories and the 10G transceiver.

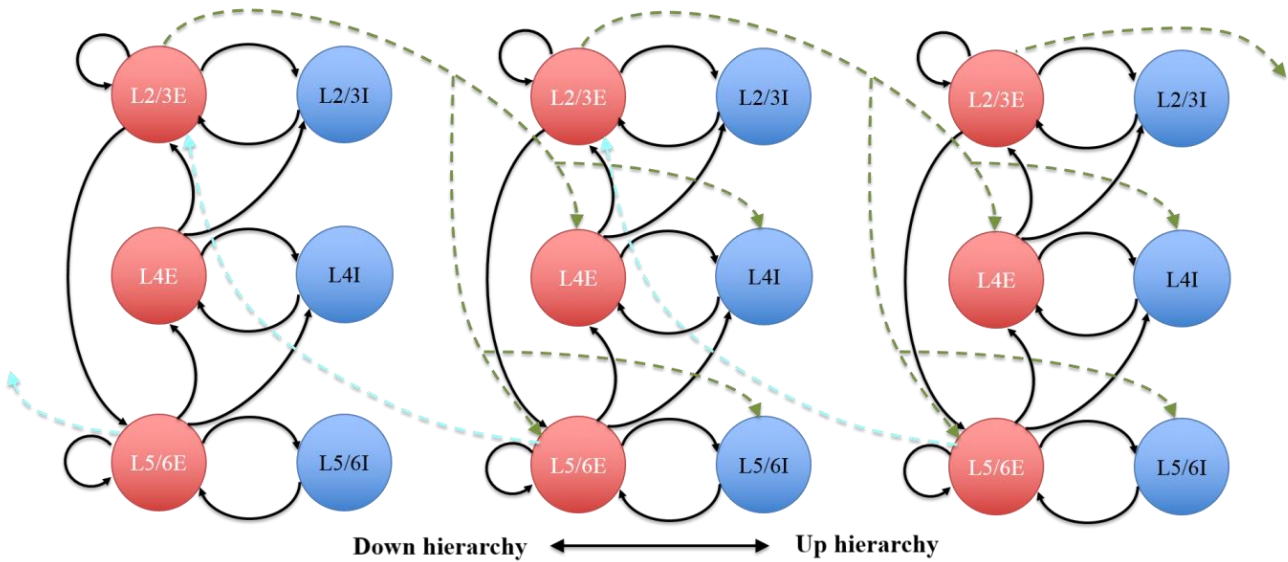

**Figure 10. The architecture of the hypercolumns.** Circles represent discrete groups. Solid and dotted arrows represent intra- and intercortical hypercolumn synaptic projections, respectively.

Along with the hardware platform, we also developed a simple application programming interface (API) in Python that is similar to the PyNN programming interface (Davison et al., 2008). This API is very similar to the high-level object-oriented interface that has been defined in the PyNN specification: it allows users to specify the parameters of neurons and connections, as well as the network structure using Python. This will enable the rapid modelling of different topologies and configurations using the cortex simulator. This API allows monitoring of the generated spikes in different hypercolumns. As future work, we plan to provide full support for PyNN scripts and incorporate interactive visualisation features on the cortex simulator.

## 3.2 Auditory cortex experiment

The results presented here focus on demonstrating the capability of our system for simulating large and structurally connected spiking neural networks in real time by implementing a simplified auditory cortex. The experiment ran the simulation in real time and the support system that generated the stimulus and recorded the results were implemented on a second DE5 board.

The human primary auditory cortex (A1) contains approximately 100 million neurons (MacGregor, 1993) and receives sensory input via the brainstem and mid-brain from the cochlea. One important aspect of the auditory cortex is its tonotopic map: neurons in auditory cortex are organised according to the frequency of sound to which they respond best from low to high along one dimension. Neurons at one end of a tonotopic map respond best to low frequencies; neurons at the other end respond best to high frequencies. This arrangement reflects the tonotopic organisation of the cochlear output, which is maintained throughout the brainstem and midbrain areas.

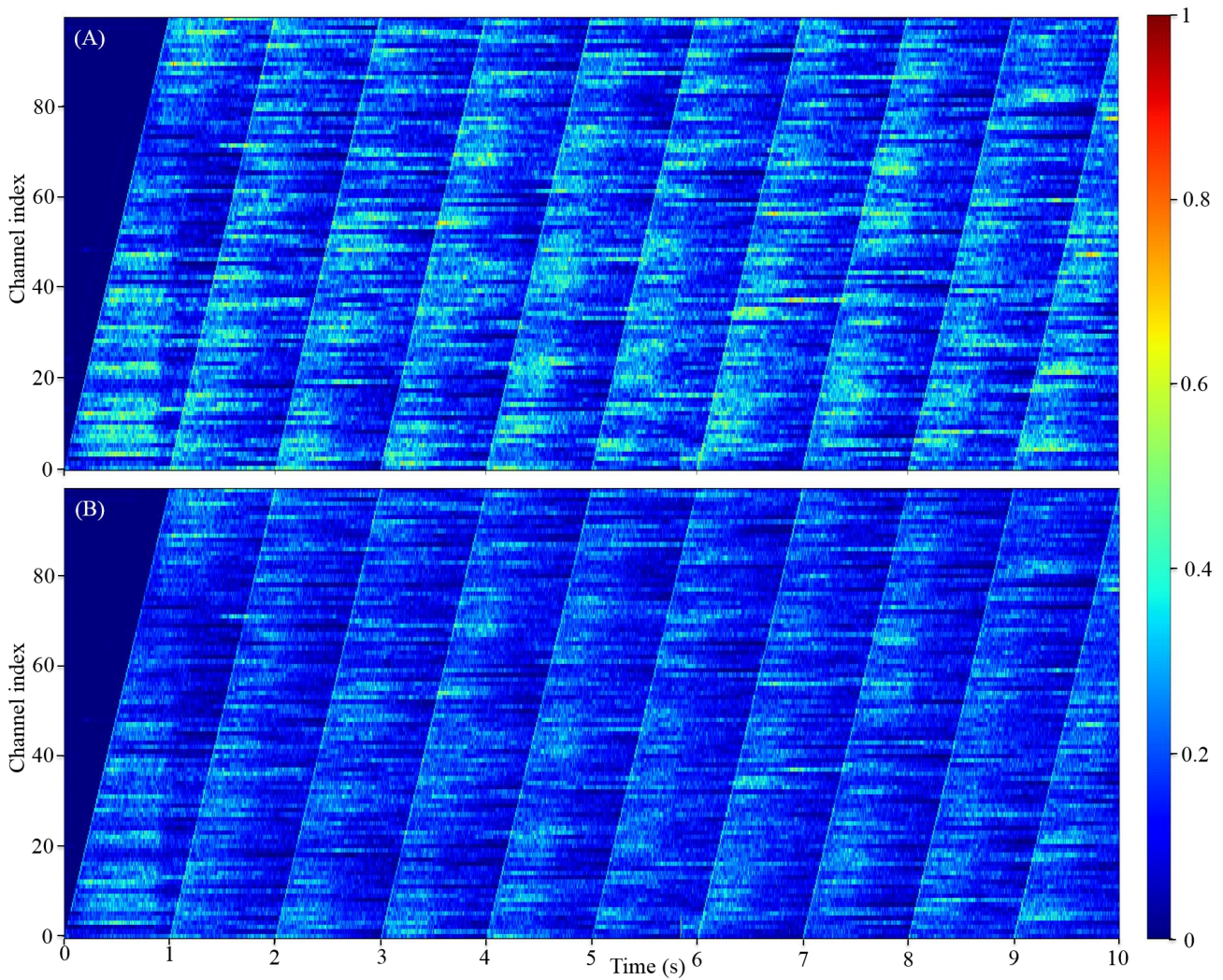

**Figure 11** | The firing activity of all the 100 channels averaged over the neurons in each channel: (A) the 800k excitatory neurons per channel; (B) the 200k inhibitory neurons per channel. The colour maps of (A) and (B) are normalised to the highest event count of the excitatory and inhibitory neurons, respectively.

The simplified auditory cortex model used in this experiment consists of 100 channels, the sensory input to which is synthetic data that emulates the output of a neuromorphic cochlea (Xu et al., 2016). This cochlea implements 100 sections (one per frequency range) in real time with incoming sound sampled at 44.1 kHz. We generated a synthetic signal with 100 channels representing the output of the cochlea in response to a pure tone sweep with exponentially increasing frequency.

Each cochlear channel is connected to a cortical population consisting of 100 hypercolumns with 100 minicolumns each. Our auditory cortex thus contains 100 × 100 = 10k hypercolumns, yielding 100 million LIF neurons. We assigned a ratio between inhibitory and excitatory neurons of 1:4. This ratio applies for each minicolumn (80 excitatory neurons and 20 inhibitory neurons). Within a minicolumn there are three groups (i.e., six neuron types) representing the different layers of neurons in cortex: L2/3, L4 and L5/6. The L2/3 group has 40 (32 excitatory and 8 inhibitory), the L4 group has 20 (16 excitatory and 4 inhibitory) and the L5/6 group has 40 (32 excitatory and 8 inhibitory) neurons, which is reasonably consistent with the number of neurons in these layers in a biological minicolumn. In the experiment, type 1-2, type 3-4 and type 5-6 implemented the excitatory and inhibitory neurons of the L2/3, L4 and L5/6, respectively. The parameters of the excitatory and inhibitory neurons are all set to the same values (see Table II). This was done for simplicity, since we are only using this experiment to verify our hardware implementation, not to reproduce any particular detailed behaviour of auditory cortical neurons. For more detailed cortical simulations, different parameters would be chosen for different neuron types.

**TABLE II. Parameters used in the BRN**

| | |
|---|---|
| $\tau_{EPSC}$ | 5.8 ms |
| $\tau_{IPSC}$ | 5.8 ms |
| $\tau_{mem}$ | 5.8 ms |
| $\tau_{rfc}$ | 3 ms |
| $g_{syn}$ | 1 |
| $g_{psc}$ | 1 |

For real time simulation, we used 176k TM minicolumns and a time slot for the axon array to access the DDR memory (between two 1k-burst operations), which was configured to be up to 200 clock cycles. Thus, the total time per update cycle is 176× (1024+200) ×5ns ≈ 1ms.

The connectivity of the hypercolumns, minicolumns and groups was also made consistent with known statistics of the anatomy of cortex (see Figure 10), which was used to verify the hierarchical communication scheme and the functions of the Parameter LUT. In our simplified auditory cortex, there are no connections between the 100 channels. Within each channel, each minicolumn is randomly connected to 24 other minicolumns: eight of them in the same hypercolumn and eight of them each in two neighbouring hypercolumns. The connections of the groups within the same hypercolumn use the intracortical connection motif, whereas the ones between hypercolumns use intercortical connectivity statistics, as shown in Figure 10. For balanced excitation-inhibition, the synaptic weights for excitatory and inhibitory connections are 0.4 and -1.0, respectively.

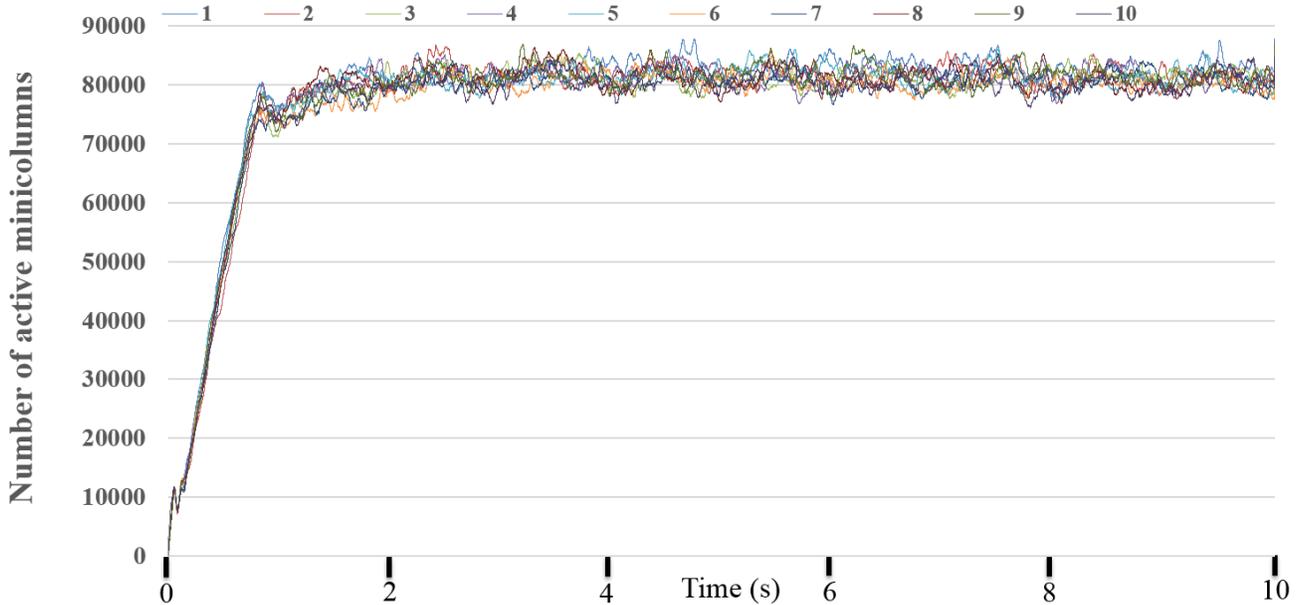

**Figure 12. The number of active minicolumns.** We run this experiment for ten times. It clearly demonstrated that the performance of the dynamic-assignment approach is stable. The mean dynamic ratio from 10 runs was about 100:8.

The sensory input is simulated using 1000 Poisson neurons (10 per channel) driven by the synthetic data that emulates the output of the cochlea. Each Poisson neuron is connected to eight randomly chosen minicolumns in a hypercolumn. The 10 Poisson neurons per channel are thus connected to 10 out of the 100 hypercolumns in that channel. The average firing rate of these Poisson neurons is approximately 10 Hz. The spikes generated by these Poisson neurons are only sent to the excitatory neurons of the L4 group, which is also consistent with known anatomy of cortex. To simulate the cochlear output of a pure tone sweep with exponentially increasing frequency, within every 10 ms, there is one set of 10 Poisson neurons generating spikes. For 100 channels, it will take 100×10ms = 1s to go through all the 100 channels. To better monitor the dynamics of the neural networks, the stimulus is repeated ten times.

Our model comprising 100 million neurons is to-date the largest spiking neural networks simulated in real-time. Since it is obviously impractical to show the firing activities of all the 100 million neurons, we chose instead to show how the firing rates of all the 100 channels change with the input signals. The results in Figure 11 show that each channel will remain silent until its input Poisson neurons fire. After that, the channel exhibits the classic asynchronous irregular network states (Brunel, 2000) for hundreds of milliseconds before falling silent or receiving the next input burst from the Poisson neurons. These results also confirm that the channels of this simplified auditory cortex are capable of emulating the heterogenous firing behaviours, due to the randomness introduced by the stochastic approach to decay in our LIF neurons and the randomness of the Poisson neurons.

To obtain the statistics of the dynamic-assignment approach, we ran the experiment for ten times while monitoring every millisecond how many TM minicolumns are being "active", i.e., at least one of its neurons is integrating or firing and/or the PSC is non-zero. Figure 12 shows the number of the active (TM) minicolumns, showing that only 80,000 out of the 1 million minicolumns are active at any given time and that this ratio remains fairly stable. It demonstrates the suitability of using the dynamic-assignment approach for sparsely connected large-scale neural networks.

We used PowerPlay, the power analysis tool provided by Altera to estimate the power dissipation of the FPGA, since a direct measurement of the FPGA's power consumption is not possible on this development kit. PowerPlay estimated the power dissipation (see Table III) from the netlist of the design, using the tool's default settings. The total power dissipation of the DDR and QDR memories are 6.6 W, which were estimated by the tools of Micron and Cypress. The total power dissipation of the whole system is ~32.4 W. Since there are 20M (TM) neurons, the power dissipation per (TM) neuron is 32.4W/20M≈1.62 μW.

**TABLE III. FPGA's power dissipation estimated by PowerPlay**

| | |
|---|---|
| Total power dissipation | 25.8 W |
| I/O power dissipation | 8.1 W |
| Core static power dissipation | 3.6 W |
| Core dynamic power dissipation | 14.1 W |
| DDR interface dynamic power dissipation | 502 mW |
| QDR interface dynamic power dissipation | 238 mW |
| Minicolumn array dynamic power dissipation | 369 mW |
| Synapse array dynamic power dissipation | 7.5 W |
| Axon array dynamic power dissipation | 582 mW |
| Parameter LUT dynamic power dissipation | 239 mW |
| Master dynamic power dissipation | 247 mW |
| Routing dynamic power dissipation* | 4.4 W |

*This item includes all the dynamic power dissipation caused by the routing

## 4. Discussion

Existing neuromorphic platforms were developed for different applications with different specifications, and different trade-offs, and are difficult to compare. Since we are mainly interested in the trade-off between the scale of the implementation, i.e., the number of neurons that the system can simulate, and the hardware cost, we will focus here on comparing to solutions that can simulate more than 100k neurons. As a result of our hardware focus, modelling and simulations of neocortex with biophysically detailed models of each individual neuron as the work studied by Markram et al. (Markram, 2015), are out of the scope of this paper and will not be addressed.

There are five large-scale neuromorphic systems: SpiNNaker, BrainScaleS, Neurogrid, HiAER-IFAT and TrueNorth. Owing to the fact that these systems were implemented with a diverse range of approaches (for diverse objectives), each of them has its own advantages and disadvantages. Table IV summarises some of the main areas for comparison. As our work was aimed at simulating large-scale spiking neural networks, of particular note are:

- Availability: Our system uses commercial off-the-shelf FPGAs, while all the other four platforms use specialised hardware that are not easily accessible to other researchers. The availability of our system will enable many more researchers to participate.

- Scalability: Our system is scalable in two ways. First, our system can be linearly scaled up using multiple FPGAs without performance loss on the back of its modular architecture and hierarchical communication scheme. Second, our system can leverage the rapid growth in FPGA technology without additional costs.

**TABLE IV. Comparison of with other large-scale neuromorphic systems**

|  | **Neurogrid** | **BrainScaleS** | **TrueNorth** | **HiAER-IFAT** | **SpiNNaker** | **This work** |
|---|---|---|---|---|---|---|
| Technology | Analogue | Analogue | Digital | Analogue | Digital | Digital |
| Feature size | 180nm | 180nm | 28nm | 130nm | 130nm | 28nm |
| Chips | 16 | 352 | 16 | 16 | 48 | 1 |
| Neurons | 1M | 200k | 16M | 1M | 768k | 20M* |
| Synapses | 4G | 40M | 4G | 1G | 768M | 4T* |
| Power | 3W | 500W | 3.2W | 40W | 80W | 32W |
| Interconnect | Tree-multicast | Hierarchical | 2D mesh-unicast | Hierarchical | 2D mesh-unicast | Hierarchical |
| Neuron model | Adaptive quadratic IF | Adaptive exponential IF | Configurable LIF | 2-compartment LIF | Programmable | 2-compartment LIF |
| Synapse model | Shared dendrite | 4-bit digital | Binary, 4 modulators | Shared conductance-based | Programmable | Shared conductance-based |
| Axon model | Axonal arbor | Speed-up Programmable delay | Four types of axonal delay | digital programmable delay | 4-bit digital programmable delay | 4-bit digital programmable delay |

*To provide a fair comparison, the dynamic assignment ratio was set to 1:1.

- Reliability: Our system is a fully digital design and it is largely insensitive to process variations and device mismatch, which are the major issues for large-scale analogue and mixed-signal designs.
- Flexibility: Dedicated hardware platforms are generally unchangeable after being fabricated. Hence, they cannot meet the requirement of rapid prototyping. Owing to the programmable nature of FPGAs, our system can be easily reconfigured for different models.

The main contributions in this paper compared to our previous work presented in (Wang et al., 2014c) are:

- Using minicolumns (including the sub-type neurons) and hypercolumns, so the parameters can be shared and therefore their number reduced so that only on-chip memory is needed to store them.
- Using structural connections between the neuron types in the minicolumns and hypercolumns, avoiding the need for the terabytes of memory that would be needed to store all the individual connections.
- Communicating spike counts instead of individual spikes; the neurons of a single type within a minicolumn all receive the same input events, so spikes do not need to be replicated individually for each neuron, reducing the required communication bandwidth.
- Improved time-multiplexing approach enabling the use of external memories for neural states.

Despite of the advantages of the FPGAs, most FPGA-based systems have been limited to behavioural tasks and failed to carry out investigations on more complicated biological plausible neural networks. This is due to the fact that an FPGA, as a digital device, does not provide massive numbers of efficient, built-in operators for highly computationally intensive functions, such as exponentials and divisions used in neuronal ion channel models. Although various designs have been presented for this challenge (Graas et al., 2004; Mak et al., 2006; Pourhaj and Teng, 2010), all the large-scale neural networks presented in the literature review were implemented with integrate-and-fire neurons or Izhikevich neuron models. In our future work, we will continue work on this system for investigating the implementation of more complex neuron models.

In addition, future developments of the design also include developing support for synaptic plasticity, e.g., Spike Timing Dependent Plasticity (STDP, Bi and Poo, 1998; Gerstner et al., 1996; Magee, 1997; Markram et al., 1997) and Spike-Timing-Dependent Delay Plasticity (STDDP, Wang et al., 2011a, 2011b, 2012, 2013a). This extension will be straightforwardly carried out by using the scheme introduced in our previous work (Wang et al., 2015), which implements a synaptic plasticity adaptor array that is separate from the neuron array. For each synapse, which remains part of the neuron, a synaptic adaptor will be connected to it when it needs to apply a certain synaptic plasticity rule. The synaptic adaptor will perform the weight/delay adaptation. This strategy enables a hardware neural network to be configured to perform multiple synaptic plasticity rules without needing to change its own structure, simply by connecting the neurons to the appropriate modules in the synaptic plasticity adaptor array. This structure was first proposed by Vogelstein and his colleagues in the IFAT project (Vogelstein et al., 2007).

Our cortex simulator uses 200k TM minicolumns to simulate 20 million to 2.6 billion LIF neurons in real time. When running at five times slower than real time, it is capable of simulating 100 million to 12.8 billion LIF neurons, which is the maximum network size on the chosen FPGA board, due to memory limitations. Larger networks could be implemented on larger FPGA boards with more external memory. A hierarchical communication scheme allows one neuron to have a fan-out of up to 200k neurons. With the advent of inexpensive FPGA boards and compute power, our cortex simulator offers an affordable and scalable tool for design, real-time simulation, and analysis of large-scale spiking neural networks.

# 5. Acknowledgement

This work was supported by the Australian Research Council Grant DP140103001. The support by the Altera university program is gratefully acknowledged.